\pdfoutput=1

\documentclass[runningheads]{llncs}

\usepackage[utf8]{inputenc}
\usepackage{graphicx}
\usepackage{mathtools}

\DeclarePairedDelimiter{\floor}{\lfloor}{\rfloor}

\usepackage{amsmath,amssymb} 
\usepackage{color}

\usepackage[table,xcdraw]{xcolor}
\usepackage[]{adjustbox}

\usepackage{color,soul}
\usepackage{subfig}
\usepackage[labelfont=bf]{caption}
\usepackage{stackengine}

\newcommand{\etal}[0]{~{\it et.~al.}}

\newcolumntype{C}[1]{>{\centering\let\newline\\\arraybackslash\hspace{0pt}}m{#1}}

\begin{document}

\title{Continual Occlusions and Optical Flow Estimation} 
\titlerunning{ContinualFlow} 



\author{Michal~Neoral \and Jan~Šochman \and Jiří~Matas}

%

\authorrunning{M. Neoral et al.} 


\institute{Center for Machine Perception, Faculty of Electrical Engineering\\Czech Technical University in Prague, Czech Republic\\
\email{\{neoramic,jan.sochman,matas\}@fel.cvut.cz}}

\maketitle

\begin{abstract}
Two optical flow estimation problems are addressed: i) occlusion estimation and handling, and ii) estimation from image sequences longer than two frames. 
The proposed ContinualFlow method estimates occlusions before flow, avoiding the use of flow corrupted by occlusions for their estimation.
We show that providing occlusion masks as an additional input to flow estimation improves the standard performance metric by more than 25\% on both KITTI and Sintel.
As a second contribution, a novel method for  incorporating information from past frames into flow estimation is introduced.
The previous frame flow serves as an input to occlusion estimation and as a prior in occluded regions, i.e. those without visual correspondences.
By continually using the previous frame flow, ContinualFlow performance improves further by 18\% on KITTI and 7\% on Sintel, achieving top performance on KITTI and Sintel.
\end{abstract}



\section{Introduction}
\label{sec:intro}
Optical flow is a two-dimensional displacement field describing the projection of scene motion between two images.
Occlusions caused by scene motion contribute to the ill-posedness of  optical flow estimation -- at occluded pixels no visual correspondences exist.
Classical non-CNN methods address this problem by using regularisation which extrapolates the flow from the surrounding non-occluded area.
Current state-of-the-art CNN algorithms for optical flow use the correlation cost volume~\cite{Dosovitskiy2015,Ilg2016,Ranjan2016,Sun2017,Meister2017,Hui2018} to estimate the most likely correspondences.
Their regularisation is only implicit and the network has to learn when to rely on the cost volume and  when to extrapolate. 
In both cases, the occluded areas are processed the same way as non-occluded ones which leads to errors in the occluded areas as well as in the nearby non-occluded regions.
 
Approaches dealing with occlusions~\cite{Meister2017,Bailer2015} usually first estimate initial forward and backward optical flows. 
Occlusions are found by a forward-backward consistency check and occlusion maps are then used for estimating of the final optical flow.
The problem here is that occlusions affect the initial flow and thus the final output.

As our first contribution, we extend a current state-of-the-art CNN optical flow method~\cite{Sun2017} by estimating the occluded areas first, {\it without estimating the flow}, and then passing the occlusion maps to the optical flow estimation network.
The correlation cost volume for flow estimation is re-used for occlusion estimation. Intuitively, the cost will be low in non-occluded areas with good correspondences and high in occluded regions.
While preserving end-to-end trainability, we accurately estimate occlusions and significantly improve the estimated flow.

Optical flow estimation over more than two frames is a problem  whose difficulty stems from the need for the pixels to be mapped to a reference coordinate system before loss evaluation.
The mapping is defined by the unknown optical flow itself.
Hence, it is difficult to apply temporal regularisation  before the flow is known.
A typical solution over three frames is to use the middle one as the reference  defining the coordinate system and to compute the forward flow to the future frame and the backward flow to the past frame and to apply regularisation to these two flows. 
Published multi-frame approaches assume various motion constraints: constant rigid motion for  three images~\cite{Wulff2017}, adaptive trajectory regularisation over five images~\cite{Volz2011}, multi-frame subspace constrains~\cite{Irani2002} and other complex motion models~\cite{Garg2013} over the whole sequence.

We avoid modelling the motion regularity explicitly and let a CNN model learn the relations of the current and previous optical flows.
The CNN is fed pairs of consecutive images together with the flow computed between the penultimate and last images.
We solve the coordinate system mapping by bilinear warp~\cite{Jaderberg2015}.
The proposed method is not limited to a fixed temporal horizon, the network uses previously estimated flows and thus, by recursion, all prior frames.

The two above-mentioned problems -- occlusion estimation and the use of multiple frames -- are related.
Since there are no correspondences in occluded areas, optical flow cannot be estimated from the cost volume and the CNN is forced to use regularisation.
Knowing the occlusions and given the previous flow, the network has prior information about the motion to be used when no correspondences are available.
So, the last estimated flow is also fed into the occlusions estimation as it is a source of information about possible occlusions.

Finally, we add a specialised refinement network~\cite{Ilg2016,Pang2017} to the proposed architecture.
It has been shown to improve fine detail accuracy of the flow, which is confirmed by our experiments.
We integrate this network with both occlusion estimation and temporal processing.

{\bf Contributions.}  We introduce integrated occlusion estimation, i.e. the algorithm does not operate on an occlusion-ignorant flow estimate,
to the state-of-the-art PWC-net~\cite{Sun2017}. 
Second, we propose a novel method that implicitly uses all previous frames for optical flow estimation. 
Finally, we add refinement blocks with additional feature map inputs leading to improved spatial resolution  of the final flow.
ContinualFlow  is state-of-the-art on several public benchmarks%
\footnote{As of the submission date, July 7, 2018.}: 1st place in Sintel~\cite{Butler2012}\footnote	{The ``Final pass'' category.} and 1st place in the KITTI'15~\cite{Menze2015} optical flow benchmark among Robust Vision Challenge (ROB) participants and 3rd over all optical flow methods\footnote{Excluding scene flow methods.} with a large margin in precision in occluded areas. Continual flow ranked 3rd in ROB~\cite{RVC2018} for the  optical flow category.


\section{Related Work}

\paragraph{\bf Occlusion estimation and occlusion handling.} 
Most optical flow methods detect occlusions as outliers of the correspondence field~\cite{Bailer2015,Gueney2016,Bailer2017} or by a consistency check on the estimated forward and backward optical flows~\cite{Sundaram2010,Chen2016}.
The optical flow is then extrapolated into the occluded areas.
The shortcoming of such approaches is that the initial flow is already adversely affected by the occlusions.
Other methods incorporate occlusion estimation directly into the energy minimisation~\cite{Xiao2006,Unger2012,Sun2014a} by truncating the data term, avoiding the problematic post-processing of already affected optical flow.
The current best performing non-CNN method~\cite{Hur2017} formulates optical flow estimation symmetrically - estimating the forward and backward flows, occlusions and dis-occlusions in a single joint optimisation.

Most of the current state-of-the-art CNN networks~\cite{Dosovitskiy2015,Ilg2016,Ranjan2016,Sun2017} do not explicitly deal with occlusions. 
The network in~\cite{Meister2017} estimates the forward and backward flows independently and uses the forward-backward consistency check to estimate the occlusions. The estimated occlusions are then used for network training only.
In LiteFlowNet~\cite{Hui2018} an occlusion probability map is a function of brightness inconsistency between the reference frame and warped target frame.
The occlusion probability map is used in a flow regularisation module.

To our best knowledge, no published CNN method estimates occlusions prior to optical flow estimation to improve the flow in the test phase.

\paragraph{\bf Using multiple frames.}
Most methods that process more than two frames impose some kind of regularisation on the flow. Murray and Buxton~\cite{Murray1987} introduced an approach that uses spatio-temporal smoothness term which regularises optical flow trajectory over multiple frames. However, the algorithm does not work well for large displacements. 
Black et al.~\cite{Black1991} extrapolate the flow from the previous frame as a starting point for the optimisation in the current frame.
In Garg et al.~\cite{Garg2010}, the motion regularisation was relaxed from several rigid motions into multi-frame subspace constraints allowing non-rigid motions.
Multi-frame subspace constraints were used in~\cite{Irani2002} over long trajectories. Its extension~\cite{Garg2013} allows more complex motions using soft constraints between frames.
An adaptive trajectory regularisation over five consecutive frames was used in~\cite{Volz2011}, where optical flow was parametrised w.r.t. the central reference frame.
Wulff\etal~\cite{Wulff2017} use super-pixel segmentation and a rigid motion assumption over triplets of images. ProFlow~\cite{Maurer2018} uses three consecutive frames, a CNN regularises  non-CNN-estimated forward ($I_t\rightarrow I_{t+1}$) and backward ($I_{t-1}\rightarrow I_t$) optical flows.

While many non-CNN algorithms use more than two frames in some form, to our best knowledge, no CNN-based method using more frames has been published.
Unlike the above-mentioned approaches, the proposed method trains the regularisation from data and does not need any hand-crafted approximations. 

\paragraph{\bf The refinement network.} The last important component added to the proposed architecture is a specialised refinement network~\cite{Ilg2016,Pang2017}.
We confirm it improves accuracy of fine details of the flow. We integrate the network with both occlusion estimation and temporal processing.

The refinement network was introduced in~\cite{Ilg2016} for optical flow estimation as a part of an architecture specialised on optical flow fine detail refinement.
The inputs to the network are the optical flow estimated by previous blocks, the brightness error of the warped image and the input images themselves.
In~\cite{Ilg2016,Pang2017}, it was shown that training the first flow estimation block and the refinement network sequentially leads to improvements in estimated optical flow.


\section{ContinualFlow}
\label{sec:proposed}
The proposed ContinualFlow method builds on the state-of-the-art  PWC-Net architecture~\cite{Sun2017}. We extend the architecture by adding i) occlusion estimation blocks and use the estimated occlusions for flow estimation, ii) an refinement network to improve fine detail accuracy, and iii)  temporal connections for utilising the previous flow for estimation of both the flow and the occlusions. Fig~\ref{fig:pwc_occl_net} shows a schematic of the PWC-Net with both the occlusions estimation blocks and temporal connections. Another diagram containing also the refinement network is shown in Fig~\ref{fig:cont_flow}.

The original PWC-Net~\cite{Sun2017} is composed of two networks: a {\it feature pyramid extractor} and a coarse-to-fine {\it optical flow decoder}.
The feature pyramid extractor takes as input two images $I_t$ and $I_{t+1}$ and encodes them into a pyramid of feature vectors $\mathcal{F}_t^s$ and $\mathcal{F}_{t+1}^s$ with gradually decreasing spatial resolution (indexed by $s$) and with increasing channel dimension. The decoder, in a coarse-to-fine manner, takes features from the corresponding resolution $s$, warps features $\mathcal{F}_{t+1}^s$ using the up-sampled flow $F_{t+1}^{s-1}$ estimated at a coarser iteration $s-1$ (if not at the coarsest resolution) and builds a correlation cost volume - a volume of feature correlations over a limited displacement range. The cost volume is then fed to the optical flow estimator, which produces the current scale optical flow $F^s$ and the process is repeated for higher resolution. We refer the reader to the original paper for further details. We are using the version with DenseNet~\cite{Huang2016} and a context network as described in the original paper.

\subsection{Occlusion Estimation}
\begin{figure*}[t]
\centering
\includegraphics[width=1.0\linewidth]{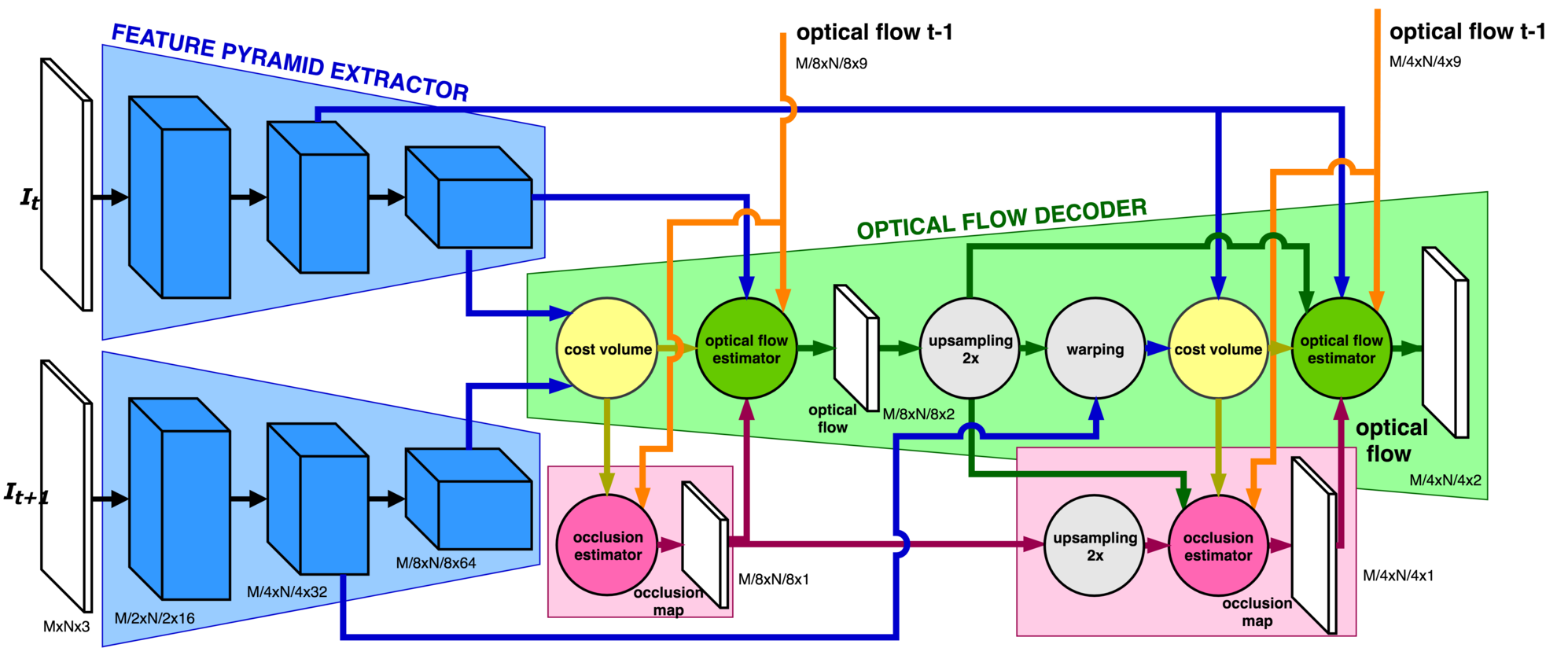}
\vspace*{1ex}
\caption{ContinualFlow - optical flow and occlusion decoder, which extends
the PWC-Net~\cite{Sun2017} flow decoder for occlusion estimation.
The feature pyramid extractor (in blue) is a convolutional network which produces a feature pyramid given an input image.
A correlation cost volume is computed on each scale from warped features from the second frame using up-sampled flow estimated at a coarser level of decoder.
The cost volume is used to estimate occlusions in occlusion estimator (in magenta). The cost volume and the occlusion map are inputs to the optical flow estimator.
For clarity, the diagram shows only three of the six levels of the ContinualFlow pyramid extractor. The output resolution is quarter of the input reference frame. Please, refer to the text for additional network details and inputs explanation.\vspace*{-1ex}}
\label{fig:pwc_occl_net}
\end{figure*}

PWC-Net and many other state-of-the-art approaches rely on the correlation cost volume for estimation of the optical flow~\cite{Dosovitskiy2015,Ilg2016,Ranjan2016,Sun2017,Meister2017}. Apart from being useful for the flow estimation, it is also indicative of possible occlusions.
Intuitively, when the cost for all displacements for some pixel is high, the pixel is likely occluded in the next frame.
In order to utilise this information, we propose to connect the occlusions estimator directly after the cost volume computation, even before any flow is estimated as shown in Fig~\ref{fig:pwc_occl_net}.
The output of the occlusions estimator is then sent to the optical flow estimator together with the cost volume itself.
This way the occlusion estimation does not rely on the imprecise flow estimation and the flow estimator benefits from the additional input.
Same as the flow estimator, the occlusions estimator works in a coarse-to-fine manner with higher resolution estimators receiving also up-sampled flow estimate from the lower resolution.

In experiments, we use an occlusion estimator with five convolutional layers with $D$, $\floor[\big]{\frac{D}{2}}$, $\floor[\big]{\frac{D}{4}}$, $\floor[\big]{\frac{D}{8}}$ and two output channels (occluded/not occluded maps), where $D=89$ in our case (the number of correlation cost volume layers + 8).
All layers use ReLU activation except for the last one, which uses soft-max.

\subsection{Refinement Network}
It was shown that a specialised refinement network which processes the output of the initial network boosts the precision of the flow estimate, especially the fine details recovery~\cite{Ilg2016,Pang2017}.
The refinement network takes several extra inputs, like the current estimate of the optical flow, image $I_{t+1}$ warped back to time step $t$ and brightness error between $I_t$ and the warped $I_{t+1}$, and produces a refined optical flow~\cite{Ilg2016}.

The refinement network used in ContinualFlow has the same architecture as the optical flow decoder, but without the DenseNet connections.
The main difference is in the network inputs.
Instead of using the input images and their warps as in~\cite{Ilg2016}, we use the features from the feature pyramid on the corresponding scale and their warps as a richer input representation.
The input error channel for these features is computed as a sum of the $L_1$ distance and structure similarity (SSIM)~\cite{Wang2004}.
We applied the refinement two times, additional refinements did not improve the accuracy in our experiments.

\subsection{ContinualFlow Estimation over Image Sequence}
\label{sec:cont_flow}
\begin{figure*}[t]
\centering
\includegraphics[width=1.0\linewidth]{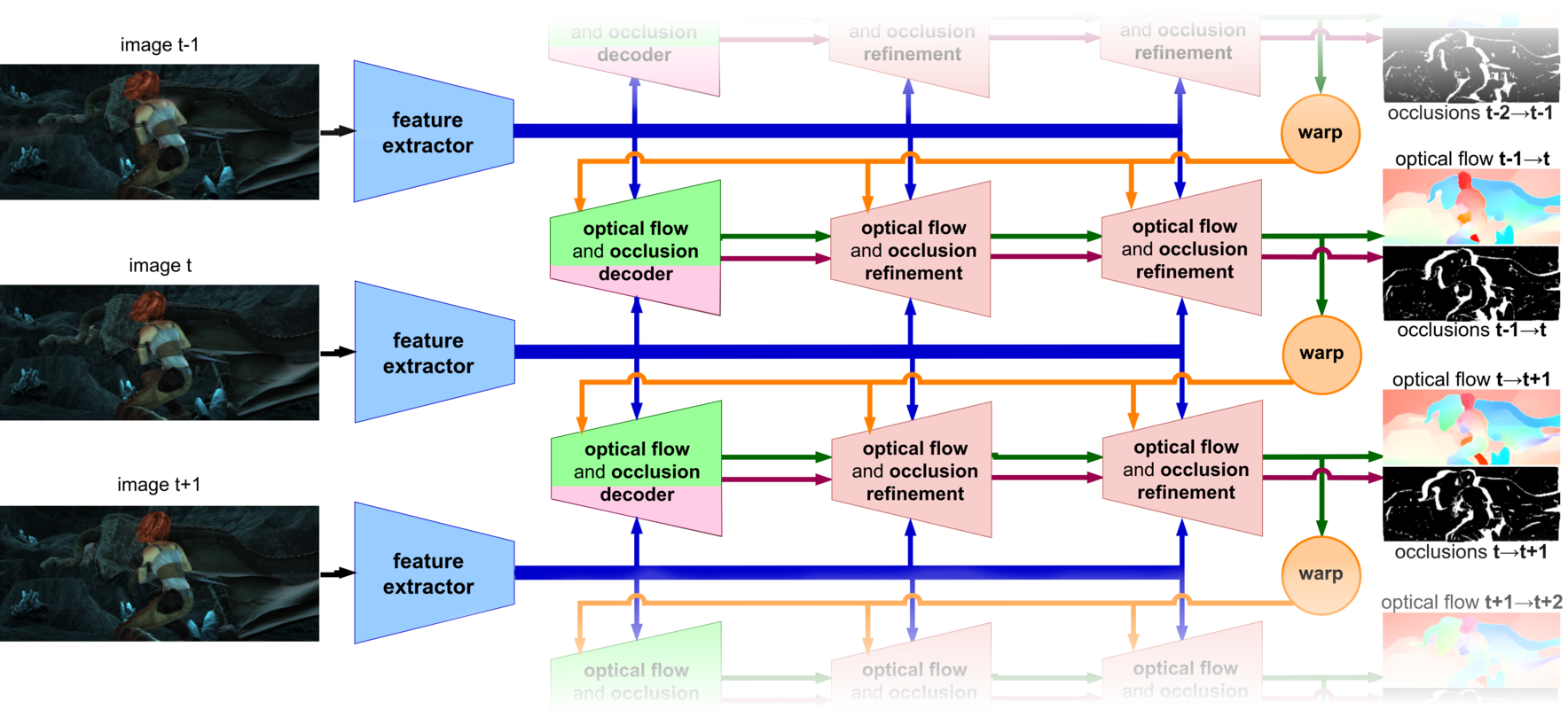}
\caption{Block diagram of ContinualFlow. Feature extractors with shared weights compute a feature pyramid from the input images. Features are input to the optical flow and occlusion decoder and the refinement blocks. The decoder estimates the optical flow and the occlusion map from the input features and from the temporal connection -- the warped optical flow from the previous time step. Optical flow and occlusion maps are finalised by the refinement blocks.}
\label{fig:cont_flow}
\end{figure*}

We use temporal connections, which give the optical flow decoder, the occlusions decoder and the refinement network an additional input: the flow estimated in the previous time step (see the orange arrows in Fig~\ref{fig:pwc_occl_net} and Fig~\ref{fig:cont_flow}).
When processing sequences longer than two frames these connections allow the network to learn typical relations between the previous and current  flows and use them in the current frame flow estimation.

However, as discussed in Sec~\ref{sec:intro}, the coordinate systems in which the two flows are expressed differ and need to be transformed onto each other in order to apply the previous flow to the correct pixels in the current time step.
Here we describe two such transformations, forward and backward warping, and we test them independently as well as in combination (concatenation of both) in Sec~\ref{sec:experiments}.

\paragraph{\bf\textbf{Forward warping transformation.}} Forward warping transforms the coordinate system from time step $t-1$ using the optical flow $F_{t-1}$ itself. The warped flow $\hat{F}_{t-1}$ is computed as
\begin{equation}
\hat{F}_{t-1}\left({\bf x} + \mbox{round}(F_{t-1}\left({\bf x}\right))\right) = F_{t-1}\left({\bf x}\right),
\end{equation}
for all pixel positions ${\bf x}$. For positions to which the flow $F_{t-1}$ maps more than once we preserve the larger of the mapped flows.
This prioritises larger motions, thus faster moving objects. Although the experiments show usefulness of this warping, the main disadvantage of this approach is that the transformation is not differentiable.
Thus, the training cannot propagate gradients through this step and relies on the shared weights only.

\paragraph{\bf\textbf{Backward warping transformation.}} Alternatively, the coordinate system could be transformed using the backward flow $B_t$ from frame $t$ to frame $t-1$.
This requires an extra evaluation of the network, but then the warping is a direct application of the differentiable spatial transformer~\cite{Jaderberg2015}.
Thus, in this case the gradients are propagated through the temporal connections during training.
A disadvantage of this approach is the computationally expensive computation of the backward flow.


\paragraph{\bf\textbf{Combining forward and backward warping.}} 
It is possible to use both warpings at the same time.
In ContinualFlow we combine forward warped previous flow, backward warped previous flow and backward flow by simply concatenating their outputs.
The only difference becomes that the previous flow input has nine channels: three times two for the flow warps and a validity masks for each warp (set to zero if the measurement is not available, e.g. at the beginning of the sequence).

\paragraph{\bf\textbf{Multi-frame sequence initialisation.}}
The network is fed a pair of input images and the previously estimated flow. 
For the first frame in the sequence, no previous flow estimation is available.
We estimate the initial optical flow between the first and second frame twice. First, we mask out the temporal connection and, in the second estimation, we use the first estimate as a temporal input.

\subsection{Training Loss}
The network is trained end-to-end with a weighted multi-task loss over the flow and occlusions estimators at all scales,
\begin{equation}
\mathcal{L} = \sum_{s=1}^S \alpha^s \mathcal{L}^{s}_F + \alpha_O \sum_{s=1}^S  \alpha^s \mathcal{L}^{s}_O\,,
\end{equation}
where $\alpha^s$ is the weight of individual scale $s$ losses and $\alpha_O$ is the occlusion estimation weight. The sums are over all $S$ spatial resolutions.
The flow estimator loss $\mathcal{L}_F$ is the same as in PWC-Net, i.e. the end-point error
\begin{equation}
\mathcal{L}^s_F = \sum_{\bf x} \gamma({\bf x}) || F^s({\bf x}) - F_{gt}^s({\bf x}) ||_2\, ,
\end{equation}
where $F^s$ is the estimated optical flow at scale $s$, $F^s_{gt}$ the corresponding ground-truth optical flow and $\gamma$ is the valid ground-truth flow mask (one for valid flow and zero otherwise). The sum is over all pixel positions.
As in~\cite{Xie2015,Caelles2017} we adopted the  weighted pixel-wise cross-entropy loss for occlusion map estimation
\begin{equation}
\begin{split}
\mathcal{L}_O^s = &- w_{noc} \sum_{{\bf x}:\, O_{gt}({\bf x}) = 1} \rho({\bf x}) \log {\rm Pr}(O({\bf x}) = 1|X)\\
&- w_{occ} \sum_{{\bf x}:\, O_{gt}(\bf x) = 0} \rho({\bf x})  \log {\rm Pr}(O({\bf x}) = 0|X)\,,
\end{split}
\end{equation}
where ${\rm Pr}(O({\bf x}) = 1|X)$ is computed using soft-max $\sigma(\cdot)$ function on the occlusion estimator output $O$, $O_{gt}$ is the ground truth occlusion map, $\rho$ the valid ground-truth occlusion mask used for masking out images without ground-truth occlusions, and $w_{occ}$ and $w_{noc}$ are the fractions of occluded and non-occluded ground truth pixels respectively.

As suggested by~\cite{Sun2017}, we modify this loss for the final fine-tuning on the most complex evaluation benchmark datasets. Here we change the $\mathcal{L}^s_F$ loss to the generalised Charbonnier loss (with $q=0.4$, $\epsilon=0.01$ as in~~\cite{Sun2017}):
\begin{equation}
\mathcal{L}^s_F = \sum_{\bf x} \gamma({\bf x}) \left( | \hat{F}^s({\bf x}) - F_{\bf gt}^s({\bf x}) | + \epsilon \right)^{q}.
\end{equation}

\section{Experiments}
\label{sec:experiments}
\paragraph{\bf{Training details.}}
The ContinualFlow network is trained using a curriculum learning approach~\cite{Bengio2009} starting from a dataset with less complex motions and increasing gradually the task complexity~\cite{Ilg2016,Sun2017}.
First, we train on FlyingChairs dataset~\cite{Dosovitskiy2015} using the training parameters introduced in~\cite{Sun2017} and following the learning rate schedule from~\cite{Ilg2016}.
We do not use rotation, scaling and translation augmentations.
Since the FlyingChairs dataset contains only two frames sequences and no occlusion ground truth, we cannot train the full ContinualFlow model with temporal connections and the occlusion map estimation.
Instead, we use it for pre-training the PWC-Net part of the ContinualFlow network. 
The network is trained for 1200k iteration and the learning rate 1e-4 is divided by 2 each 200k iteration, starting from 400k.
Images in a batch of size eight are randomly cropped to $448\times 384$\,px.

Next, the all parts of the ContinualFlow network are trained on the FlyingThings dataset~\cite{Mayer2016}.
Since occlusion maps were not available for this dataset, we computed them using the available backward and forward ground truth flows and the object segmentation masks.
The mask $O_t({\bf x})$ is set to ``occluded'' for  pixel $x$ when the object labels $L_t({\bf x})$ and $L_{t+1}(F_{gt}({\bf x}))$ differ or the bi-directional consistency between backward and forward flows differs by more than one pixel.
The network is trained for 500k iteration and the learning rate, set to 1e-4 for the first 200k iterations, is divided by 2 at that point and after 100k iterations.
First, we train the network without the refinement.
Then, only the refinement is trained while all other weights are fixed.
Images in the batch of size four are randomly cropped to $768\times 384$\,px.
After cropping, optical flow pointing out of the frame is labelled as occluded.

Finally, the ContinualFlow is trained on data from six datasets: Driving~\cite{Mayer2016}, KITTI'15~\cite{Menze2015}, VirtualKITTI~\cite{Gaidon2016}, Sintel~\cite{Butler2012}, HD1K~\cite{Kondermann2016} and the FlyingChairs small motions dataset~\cite{Ilg2016}. 
These datasets, except for FlyingChairs, contain sequences longer than two frames and are suitable for the training of temporal connections.
We used the first image twice for the FlyingChairs dataset to obtain the same batch size for all input data, the loss on the estimate of the (zero) flow $F^{0,1}$ is not used.
Dense occlusion maps are available only for the Sintel and Driving datasets. We set occlusion estimation loss to zero on the rest.
The network is fine-tuned for 500k iteration and the learning rate, set to 1e-5 for the first 200k interactions, is divided by 2 at that point and after 100k iterations.
Images in batches of size four are randomly cropped to $768\times 320$\,px.
We sample images from all datasets uniformly.

We set weights for individual scales as in~\cite{Sun2017}.
Maximal displacement in the cost volume is set to four.
The same scale weights are set to train the refinement network and for the occlusion map estimation. The occlusion estimation weight $\alpha_{O}$ is set to 0.1. 
All experiments are trained with the ADAM optimiser~\cite{Kingma2014} and 0.0004 weight decay.
All parts of the network are implemented in TensorFlow.

The ContinualFlow training has the same three phases as training of PWC-Net.
Only when training the refinement network separately, there is an additional phase which updates only the refinement parameters as mentioned above.
ContinualFlow without the refinement network has 9.6M parameters, 0.8M more than the PWC-Net.
The refinement network adds 5.0M parameters, it is based on the PWC-Net-small architecture.
ContinualFlow runs at 8 FPS on KITTI-resolution of 1240x375\,px.

In the following, we focus on the Robust Vision Challenge~\cite{RVC2018}, where one trained model with the same parameters has to be evaluated on four individual benchmarks~\cite{Menze2015,Butler2012,Kondermann2016,Baker2011} instead of fine-tuning for each particular dataset independently.

\begin{figure*}[t]
\centering
\includegraphics[width=1.0\linewidth]{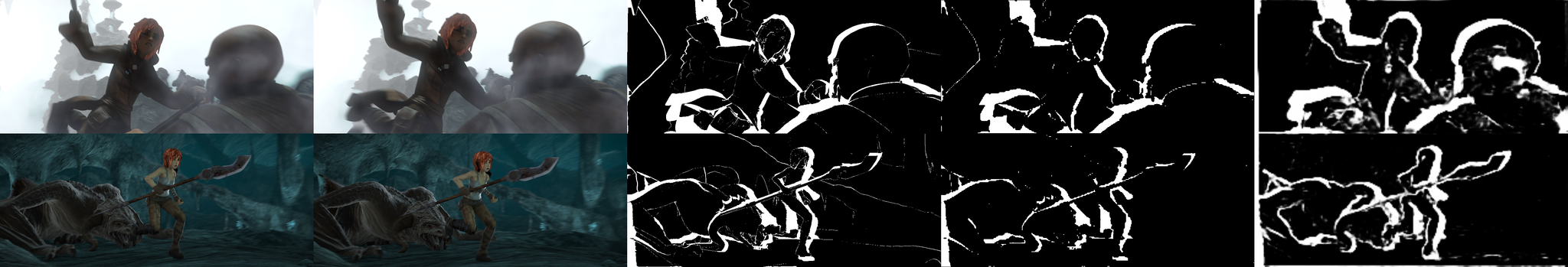}
\vspace{-1em}
\begin{tabular}{C{.192\linewidth}C{.192\linewidth}C{.192\linewidth}C{.192\linewidth}C{.192\linewidth}}
\scriptsize image $t$ & \scriptsize image $t$+1 & \scriptsize GT & \scriptsize 1/4 resolution GT & \scriptsize estimation
\end{tabular}
\caption{Example estimated occlusion maps on the Sintel (final) dataset, our validation split. ContinualFlow estimates occlusions up to quarter resolution.\vspace*{-2ex}}
\label{fig:occl_sintel}
\end{figure*}

\subsection{Ablation Study}
In this section, we experimentally evaluate the individual contributions and design choices for the ContinualFlow network trained on FlyingChairs~\cite{Dosovitskiy2015} and fine-tuned on FlyingThings~\cite{Mayer2016} as described above. 
Below, the term {\it baseline} refers to our TensorFlow implementation of PWC-Net.
Unlike the PWC-Net settings~\cite{Sun2017}, we trained  the network without rotation, scaling and translation augmentation of input frames.



\paragraph{\bf\textbf{Occlusion map learning.}} 
Table~\ref{tab:ablation_study} shows the results of optical flow estimation with and without occlusion learning.
Temporal connections are not used. Application of the occlusion map estimator improves performance on all tested datasets not only in occluded regions but also in all non-occluded regions.
Fig~\ref{fig:occl_sintel} shows example estimated occlusion maps.

\paragraph{\bf\textbf{The specialised refinement block}} improves results of  the estimated optical flow as is shown in~\cite{Ilg2016}.
Table~\ref{tab:ablation_study} compares the optical flow estimation with and without the refinement block. No temporal connections are used.
The refinement block improves the estimated optical flow, especially in occluded areas.

\paragraph{\bf\textbf{Influence of the coordinate warping methods.}}
We evaluated the three approaches for warping the previous flow estimate  introduced in section~\ref{sec:cont_flow}. Results for individual datasets are shown in Table~\ref{tab:ablation_study}. 
Forward warping $W_f$ is beneficial for the KITTI dataset~\cite{Menze2015} and the Sintel Clean dataset~\cite{Butler2012}, while backward warping $W_b$ is more suitable for the complex Sintel Final sequences.
The combination of both, $W_{bf}$,  is the most accurate on FlyingThings sequences~\cite{Mayer2016}.
%
%
All evaluated variants use the occlusion estimator in the decoder and no refinement.

\paragraph{\bf\textbf{Temporal connection placement.}}
%
%
We experimented with passing the warped optical flow from previous frame to different network components, thus creating different temporal connections.
In one variant, only the refinement network received the previous frame flow estimates. In another variant, all temporal connections as depicted in Fig~\ref{fig:cont_flow} were used.
Table~\ref{tab:ablation_study} shows how feeding these connections with different warpings influences the estimated flow.
The best results were obtained with temporal connections leading into both the decoder and refinement networks and the combination of forward and backward warpings.

\paragraph{\bf\textbf{Number of refinement blocks.}}
Table~\ref{tab:ablation_study} shows results for 1, 2, 3 and 5 stacked refinement networks.
Stacking more than two refinement networks is not beneficial.
Thus the final network architecture contains only two refinements.
All evaluated variants use the occlusion estimator and warps the previously estimated optical flow using both warping methods in the first part of the network and the refinement.

\paragraph{\bf\textbf{Multi-frame sequences initialisation}}
For the first frame in the sequence, no previous frame flow estimate is available to be passed to the temporal connections.
Unsurprisingly, the estimation on the first frame is usually slightly worse than at the consecutive frames.
We tested two initialisations of the first frame flow estimation: (i) no flow (zero displacements) instead of the previously estimated optical flow and, (ii) a two-pass initial estimation of the currently estimated optical flow as described in Section~\ref{sec:cont_flow}.
We evaluated both approaches for an increased length of the sequence on different datasets.
As shown in Table~\ref{tab:ablation_study}, the two-pass initialisation leads to quicker convergence and is most beneficial for the first optical flow estimation in the sequence.

\begin{table}[]
\caption{\textbf{Ablation study of ContinualFlow}.
The leftmost column codes the experiment configurations: occlusion estimator (+OC); refinement network (+R); temporal connection with forward warping ($\rm W_f$), backward warping ($\rm W_b$) and both warping methods ($\rm {W}_{bf}$); previous flow input in the refinement ($\rm RW_{x}$); and two pass (2pass) initialisation of the first frame of the sequence N frames long. Performance measure are the KITTI 3-pixel error metric (column Fl) and the end-point error (in pixels, all other columns) for background (bg), foreground (fg), occluded (occ), non-occluded (noc) and all (all) pixels.
%
%
The best performance in bold.
All models trained on FlyingChairs and fine-tuned on FlyingThings.
See section~\ref{sec:proposed} for details.\\}
\label{tab:ablation_study}
\centering
\begin{adjustbox}{max width=\textwidth}
\begin{tabular}{l|ccccc|cc|cc|ccc|ccc}
\textbf{} & \multicolumn{5}{c|}{FlyingThings} & \multicolumn{2}{c|}{KITTI'15 noc} & \multicolumn{2}{c|}{KITTI'15 occ} & \multicolumn{3}{c|}{Sintel Clean} & \multicolumn{3}{c}{Sintel Final} \\
 & all & occ-bg & occ-fg & noc-bg & noc-fg & Fl-all & all & Fl-all & all & all & occ & noc & all & occ & noc  \\ \hline
\multicolumn{2}{|l}{common: baseline} & \multicolumn{12}{c}{\textbf{Occlusion map learning}} & \multicolumn{2}{l|}{} \\ \hline
\multicolumn{1}{|l|}{} & 22.79 & 25.31 & 53.88 & 10.64 & 26.78 & 37.73 & 7.82 & 43.56 & 14.16 & 3.45 & 9.29 & 2.38 & 5.36 & 12.03 & \multicolumn{1}{|l|}{4.17} \\
\multicolumn{1}{|l|}{+OC} & \textbf{18.01} & \textbf{18.27} & \textbf{47.53} & \textbf{7.10} & \textbf{20.13} & \textbf{23.98} & \textbf{5.22} & \textbf{31.12} & \textbf{10.60} & \textbf{2.45} & \textbf{7.46} & \textbf{1.53} & \textbf{4.02} & \textbf{9.99} & \multicolumn{1}{|l|}{\textbf{2.91}} \\ \hline \hline
\multicolumn{2}{|l}{common: baseline+OC} & \multicolumn{12}{c}{\textbf{The specialised refinement block}} & \multicolumn{2}{l|}{} \\ \hline
\multicolumn{1}{|l|}{} & 18.01 & 18.27 & 47.53 & \textbf{7.10} & \textbf{20.13} & 23.98 & 5.22 & 31.12 & 10.60 & 2.45 & 7.46 & 1.53 & 4.02 & 9.99 & \multicolumn{1}{|l|}{2.91} \\
\multicolumn{1}{|l|}{+R} & \textbf{17.80} & \textbf{17.49} & \textbf{45.90} & 7.31 & 21.46 & \textbf{21.14} & \textbf{4.78} & \textbf{28.61} & \textbf{9.83} & \textbf{2.30} & \textbf{7.11} & \textbf{1.42} & \textbf{3.87} & \textbf{9.68} & \multicolumn{1}{|l|}{\textbf{2.76}} \\ \hline \hline
\multicolumn{2}{|l}{common: baseline+OC} & \multicolumn{12}{c}{\textbf{Influence of coordinate warping methods}} & \multicolumn{2}{l|}{} \\ \hline
\multicolumn{1}{|l|}{} & 18.01 & 18.27 & 47.53 & 7.10 & 20.13 & 23.98 & 5.22 & 31.12 & 10.60 & 2.45 & 7.46 & 1.53 & 4.02 & 9.99 & \multicolumn{1}{|l|}{2.91} \\
\multicolumn{1}{|l|}{+$\rm W_f$} & 14.90 & 14.89 & 38.75 & 6.55 & \textbf{16.69} & \textbf{20.78} & \textbf{4.13} & \textbf{27.85} & \textbf{8.28} & \textbf{2.18} & \textbf{6.67} & \textbf{1.37} & 4.04 & 9.48 & \multicolumn{1}{|l|}{3.03} \\
\multicolumn{1}{|l|}{+$\rm W_b$} & 16.33 & 17.10 & 39.68 & 6.49 & 20.72 & 26.52 & 4.56 & 33.80 & 10.64 & 2.58 & 7.49 & 1.70 & \textbf{3.79} & 9.27 & \multicolumn{1}{|l|}{\textbf{2.80}} \\
\multicolumn{1}{|l|}{+$\rm W_{bf}$} & \textbf{14.64} & \textbf{14.84} & \textbf{36.05} & \textbf{6.10} & 17.65 & 23.64 & 4.56 & 30.92 & 9.46 & 2.36 & 6.79 & 1.59 & 3.81 & \textbf{8.97} & \multicolumn{1}{|l|}{2.87} \\ \hline \hline
\multicolumn{2}{|l}{common: baseline+OC} & \multicolumn{12}{c}{\textbf{Temporal connection placement}} & \multicolumn{2}{l|}{} \\ \hline
%
%
\multicolumn{1}{|l|}{+$\rm RW_f$} & 16.10 & 15.87 & 38.71 & 6.76 & 19.01 & 23.11 & 4.88 & 30.35 & 9.82 & 2.27 & 6.89 & \textbf{1.45} & 3.92 & 9.34 & \multicolumn{1}{|l|}{2.90} \\
\multicolumn{1}{|l|}{+$\rm RW_{bf}$} & 14.90 & 15.35 & 37.55 & \textbf{5.78} & 17.69 & 24.54 & 4.84 & 32.18 & 10.12 & 2.35 & 6.93 & 1.54 & \textbf{3.55} & \textbf{8.62} & \multicolumn{1}{|l|}{\textbf{2.65}} \\
\multicolumn{1}{|l|}{+$\rm W_{bf}$} & 14.64 & 14.84 & 36.05 & 6.10 & 17.65 & 23.64 & 4.56 & 30.92 & 9.46 & 2.36 & 6.79 & 1.59 & 3.81 & 8.97 & \multicolumn{1}{|l|}{2.87} \\
\multicolumn{1}{|l|}{+$\rm W_{bf}$+ $\rm RW_{bf}$}  & \textbf{14.28} & \textbf{14.24} & \textbf{35.58} & 5.82 & \textbf{17.56} & \textbf{21.72} & \textbf{4.41} & \textbf{29.48} & \textbf{9.33} & \textbf{2.26} & \textbf{6.66} & 1.49 & 3.70 & 8.81 & \multicolumn{1}{|l|}{2.76} \\ \hline \hline
\multicolumn{2}{|l}{common: baseline+OC+$\rm W_{bf}$} & \multicolumn{12}{c}{\textbf{Number of refinement blocks}} & \multicolumn{2}{l|}{} \\ \hline
\multicolumn{1}{|l|}{+1x$\rm RW_{bf}$} & 14.28 & 14.24 & 35.58 & 5.82 & \textbf{17.56} & \textbf{21.72} & \textbf{4.41} & \textbf{29.48} & \textbf{9.33} & \textbf{2.26} & \textbf{6.71} & \textbf{1.47} & \textbf{3.76} & \textbf{8.93} & \multicolumn{1}{|l|}{2.80} \\
\multicolumn{1}{|l|}{+2x$\rm RW_{bf}$} & \textbf{14.26} & \textbf{14.13} & \textbf{35.60} & 5.78 & 17.62 & 21.77 & 4.45 & 29.62 & 9.35 & \textbf{2.26} & 6.72 & \textbf{1.47} & \textbf{3.76} & 8.96 & \multicolumn{1}{|l|}{\textbf{2.79}} \\
\multicolumn{1}{|l|}{+3x$\rm RW_{bf}$} & 14.30 & \textbf{14.13} & 35.71 & \textbf{5.75} & 17.77 & 21.98 & 4.50 & 29.86 & 9.40 & \textbf{2.26} & 6.74 & \textbf{1.47} & 3.77 & 8.99 & \multicolumn{1}{|l|}{2.80} \\
\multicolumn{1}{|l|}{+5x$\rm RW_{bf}$} & 14.43 & 14.24 & 36.16 & \textbf{5.75} & 17.93 & 22.48 & 4.58 & 30.35 & 9.49 & 2.28 & 6.80 & 1.48 & 3.80 & 9.03 & \multicolumn{1}{|l|}{2.83} \\ \hline \hline 
\multicolumn{3}{|l}{common: baseline+OC+$\rm W_{bf}$+$\rm RW_{bf}$} & \multicolumn{10}{c}{\textbf{Multi-frame sequence initialisation}} & \multicolumn{3}{l|}{} \\ \hline
\multicolumn{1}{|l|}{2 frames}& - & - & - & - & - & 25.08 & 5.50 & 32.59 & 11.56 & 2.48 & 7.72 & 1.48 & 3.84 & 9.64 & \multicolumn{1}{|l|}{2.75} \\
\multicolumn{1}{|l|}{2 frames+2pass}& - & - & - & - & - &  \textbf{23.06} & \textbf{5.03} & \textbf{30.92} & \textbf{11.00} & \textbf{2.41} & \textbf{7.60} & \textbf{1.41} & \textbf{3.74} & \textbf{9.48} & \multicolumn{1}{|l|}{\textbf{2.66}} \\ \hline
\multicolumn{1}{|l|}{3 frames}& - & - & - & - & - &  21.72 & 4.41 & 29.48 & 9.33 & 2.26 & 6.71 & \textbf{1.47} & 3.76 & 8.93 & \multicolumn{1}{|l|}{2.80} \\
\multicolumn{1}{|l|}{3 frames+2pass}& - & - & - & - & - &  \textbf{21.65} & \textbf{4.36} & \textbf{29.42} & \textbf{9.23} & \textbf{2.26} & \textbf{6.71} & 1.48 & \textbf{3.73} & \textbf{8.92} & \multicolumn{1}{|l|}{\textbf{2.76}} \\ \hline
 \multicolumn{1}{|l|}{4 frames}& - & - & - & - & - &  \textbf{21.53} & 4.30 & \textbf{29.32} & 9.05 & \textbf{2.23} & 6.59 & 1.46 & 3.75 & 8.83 & \multicolumn{1}{|l|}{2.82} \\
 \multicolumn{1}{|l|}{4 frames+2pass}& - & - & - & - & - &  21.54 & \textbf{4.30} & 29.33 & \textbf{9.02} & 2.24 & \textbf{6.59} & \textbf{1.46} & \textbf{3.73} & \textbf{8.80} & \multicolumn{1}{|l|}{\textbf{2.80}} \\ \hline
 \multicolumn{1}{|l|}{5 frames}& - & - & - & - & - &  \textbf{21.48} & \textbf{4.25} & \textbf{29.27} & \textbf{8.92} & \textbf{2.21} & \textbf{6.51} & \textbf{1.45} & 3.80 & 8.85 & \multicolumn{1}{|l|}{2.87} \\
 \multicolumn{1}{|l|}{5 frames+2pass}& - & - & - & - & - &  \textbf{21.48} & \textbf{4.25} & 29.28 & \textbf{8.92} & \textbf{2.21} & 6.52 & 1.46 & \textbf{3.79} & \textbf{8.83} & \multicolumn{1}{|l|}{\textbf{2.86}} \\ \hline
 \multicolumn{1}{|l|}{10 frames}& - & - & - & - & - & 21.49 & \textbf{4.24} & 29.28 & 8.90 & - & - & - & - & - & \multicolumn{1}{|l|}{-} \\
 \multicolumn{1}{|l|}{10 frames+2pass}& - & - & - & - & - & \textbf{21.48} & \textbf{4.24} & \textbf{29.27} & \textbf{8.89} & - & - & - & - & - & \multicolumn{1}{|l|}{-} \\\hline
\end{tabular}
\end{adjustbox}
\end{table}

\subsection{Comparison with State of the Art}
We start by noting that a single model was used for all benchmarks without further fine-tuning to individual datasets.
We were not able to evaluate occlusions on public benchmarks since there is no benchmark available for occlusion map estimation.
ContinualFlow achieves recall 0.87 and F1-score 0.83 for the validation split of FlyingThings~\cite{Mayer2016} and recall 0.72 and F1-score 0.48 for Sintel~\cite{Butler2012}. Examples of estimated occlusion maps are shown in Fig~\ref{fig:occl_sintel}.

\paragraph{\bf\textbf{KITTI'15}} optical flow benchmark~\cite{Menze2015}
results are reported in Table~\ref{tab:comparison_kitti}.
Fl refers to the KITTI evaluation metric -- the percentage of pixels with end-point-error greater than 3\,px.
Our method ranked first among methods participating in the Robust Vision Challenge (ROB) and third for all optical flow estimation methods with score 10.03\% on all evaluated pixels.
We are interested in ROB Challenge since methods outside ROB fine-tune on each particular dataset, resulting in over-fitting, which we wanted to avoid.


\begin{figure*}[t]
\centering
\subfloat{{\stackunder{\includegraphics[width=2.4cm]{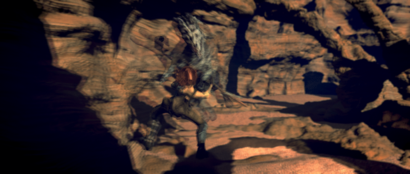} }{ }}}%
\subfloat{{\stackunder{\includegraphics[width=2.4cm]{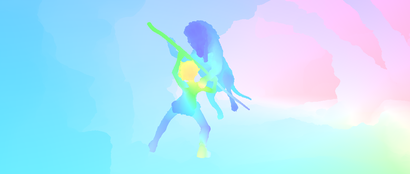} }{ }}}%
\subfloat{{\stackunder{\includegraphics[width=2.4cm]{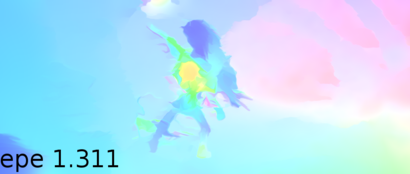} }{ }}}%
\subfloat{{\stackunder{\includegraphics[width=2.4cm]{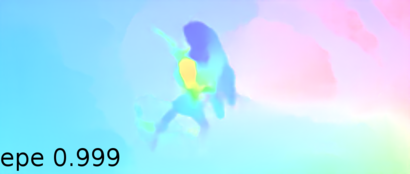} }{ }}}%
\subfloat{{\stackunder{\includegraphics[width=2.4cm]{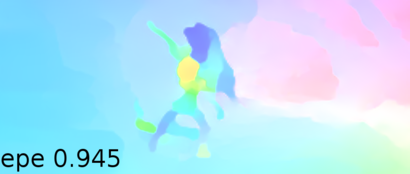} }{ }}}%
\\\vspace{-0.4cm}%
\subfloat{{\stackunder{\includegraphics[width=2.4cm]{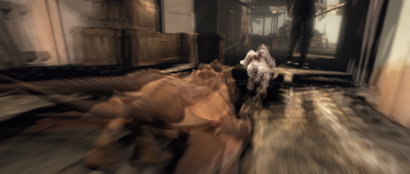} }{ }}}%
\subfloat{{\stackunder{\includegraphics[width=2.4cm]{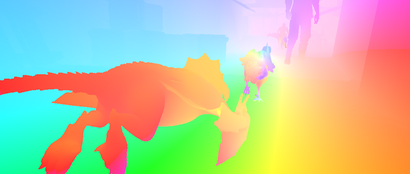} }{ }}}%
\subfloat{{\stackunder{\includegraphics[width=2.4cm]{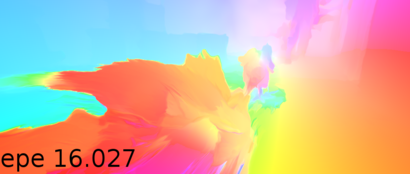} }{ }}}%
\subfloat{{\stackunder{\includegraphics[width=2.4cm]{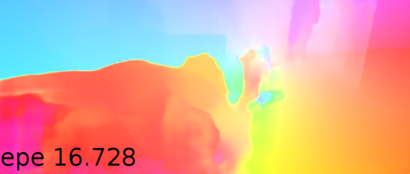} }{ }}}%
\subfloat{{\stackunder{\includegraphics[width=2.4cm]{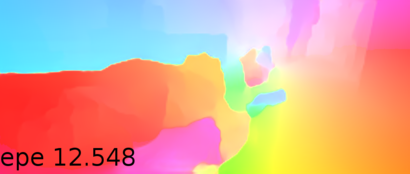} }{ }}}%
\\\vspace{-0.4cm}%
\subfloat{{\stackunder{\includegraphics[width=2.4cm]{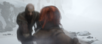}}{\scriptsize image~overlay}}}
~\,\subfloat{{\stackunder{\includegraphics[width=2.4cm]{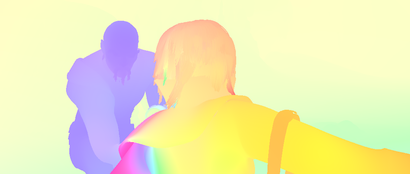}}{\scriptsize ground~truth}}}
~\subfloat{{\stackunder{\includegraphics[width=2.4cm]{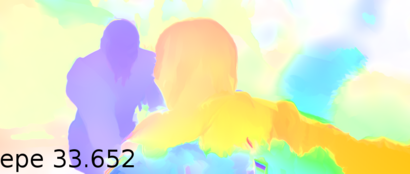}}{\scriptsize ProFlow\_ROB~\cite{Maurer2018}}}}
~\,\subfloat{{\stackunder{\includegraphics[width=2.4cm]{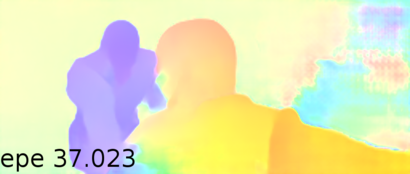}}{\scriptsize PWC-Net\_ROB~\cite{Sun2017}}}}
\,\subfloat{{\stackunder{\includegraphics[width=2.4cm]{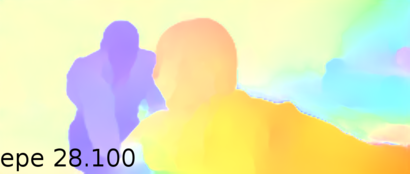}}{\scriptsize ContinualFlow\_ROB}}}
\caption{Example results on Sintel Final-pass for ContinualFlow closest competitors in the Robust Vision Challenge. End-point-error for each method is shown for particular scenes.}
\label{fig:sintel_comparison}
\end{figure*}

\begin{table}[]
\caption{KITTI'15 optical flow benchmark results of Robust Vision Challenge participants as of June 7, 2018.
Performance measured by the KITTI 3-pixel error metric (column Fl) and the end-point error (in pixels, all other columns) for background (bg), foreground (fg), occluded (occ), non-occluded (noc) and all (all) pixels.
The best results in bold.
Anonymous entries in time of paper submission are marked {[}anon{]}.
Methods are sorted according to Fl-all, the default ranking for KITTI.}
\label{tab:comparison_kitti}
\centering
\begin{adjustbox}{max width=\textwidth}
\begin{tabular}{l|rrr|rrr}
 & \multicolumn{3}{c|}{KITTI'15 occ (\%)} & \multicolumn{3}{c}{KITTI'15 noc (\%)} \\
Fl (\%) & bg & fg & all & bg & fg & all \\ \hline
\textbf{ContinualFlow\_ROB} & \textbf{8.54} & 17.48 & \textbf{10.03} & \textbf{5.90} & 14.99 & 7.55 \\ 
LFNet\_ROB {[}anon{]} & 11.18 & 10.20 & 11.01 & 6.14 & 6.87 & \textbf{6.27} \\
PWC-Net\_ROB~\cite{Sun2017} & 11.22 & 13.69 & 11.63 & 7.12 & 10.29 & 7.69 \\
ProFlow\_ROB~\cite{Maurer2018} & 14.15 & 21.82 & 15.42 & 8.44 & 17.90 & 10.15 \\
FF++\_ROB~\cite{Schuster2018a} & 15.32 & 19.27 & 15.97 & 7.82 & 15.33 & 9.18 \\
ResPWCR\_ROB {[}anon{]} & 16.63 & 16.18 & 16.55 & 10.10 & 12.23 & 10.49 \\
AugFNG\_ROB {[}anon{]} & 19.77 & \textbf{9.95} & 18.14 & 13.75 & \textbf{6.71} & 12.47 \\
DMF\_ROB~\cite{Weinzaepfel2013} & ~30.74 & ~30.07 & ~30.63 & ~19.32 & ~25.60 & ~20.46 \\
\end{tabular}
\end{adjustbox}
\end{table}

\paragraph{\bf\textbf{Sintel.}}
Fig~\ref{fig:sintel_comparison} shows visual comparison with the closest competitors. Results of ROB participants  on the Sintel dataset are reported in Table~\ref{tab:comparison_sintel}. ContinualFlow ranked first on Sintel Final for the all pixels end-point-error evaluation. As we are focused on occlusion estimation and handling, we point out that ContinualFlow achieves best results for estimation in occluded areas with significant margin.

\begin{table}[]
\caption{Sintel benchmark results for Robust Vision Challenge participants.
Performance measured the end-point error (EPE, in pixels) for matched (noc), unmatched (occ) and all (all) pixels.
The best results in bold.
Anonymous entries marked {[}anon{]}.
Methods are sorted by EPE all, the default ranking for Sintel.
}
\label{tab:comparison_sintel}
\centering
\begin{adjustbox}{max width=\textwidth}
\begin{tabular}{l|rrr|rrr}
 & \multicolumn{3}{c|}{Sintel Final} & \multicolumn{3}{c}{Sintel Clean} \\
 & all & noc & occ & all & noc & occ \\ \hline
\textbf{ContinualFlow\_ROB} & \textbf{4.528} & 2.723 & \textbf{19.248} & 3.341 & 1.752 & \textbf{16.292}\\
PWC-Net\_ROB~\cite{Sun2017} & 4.903 & \textbf{2.454} & 24.878 & 3.897 & 1.726 & 21.637 \\
ProFlow\_ROB~\cite{Maurer2018} & 5.015 & 2.659 & 24.192 & \textbf{2.709} & \textbf{1.013} & 16.549 \\
AugFNG\_ROB {[}anon{]} & 5.500 & 2.978 & 26.052 & 3.606 & 1.603 & 19.939 \\
LFNet\_ROB {[}anon{]} & 5.966 & 3.278 & 27.893 & 4.815 & 2.333 & 25.065 \\
FF++\_ROB~\cite{Schuster2018a} & 6.496 & 2.990 & 35.057 & 3.953 & 1.148 & 26.836 \\
ResPWCR\_ROB {[}anon{]} & 6.530 & 3.849 & 28.371 & 5.674 & 3.138 & 26.380 \\
DMF\_ROB~\cite{Weinzaepfel2013} & 7.475 & 3.575 & 39.245 & 5.368 & 1.742 & 34.899 \\
\end{tabular}
\end{adjustbox}
\end{table}




\paragraph{\bf\textbf{Robust Vision Challenge.}}
A snapshot of the leaderboard\footnote{As of July 7, 2018.} of optical flow Robust Vision Challenge~\cite{RVC2018} is shown in Table~\ref{tab:comparison_rob}.
ContinualFlow is built on our implementation of PWC-Net~\cite{Sun2017}. 
While ContinualFlow did not achieve a better results in the ROB than the original PWC-Net, the experiments show that our contributions outperform the results of our baseline. 

The source code for PWC-Net was released by the authors just days before the ACCV submission deadline, so
a direct comparison was possible only through ROB vision challenge submissions, which are limited in number by the challenge rules. 
We did our best to follow the paper regarding the architecture, parameters and training. 
Later, when analysing the results, we found two main differences: 
i) Due to implementation issues, we omitted rotation and scaling data augmentations, which in retrospect could harm the performance significantly as suggested in~\cite{Mayer2018}.
ii) Our implementation is in Tensorflow whereas the original implementation is in Caffe, so some of the suggested training parameter values may need to be fine-tuned for this framework. 
Still, the ablation study clearly shows the impact and significance of the novelties (occlusion estimation, feeding the previous estimate of optical flow as input).

\begin{table}[tb]
\caption{Robust Vision Challenge.
Performance measured by ranking of all metrics in individual datasets.
The best results in bold.
Anonymous entries marked {[}anon{]}.
Methods are sorted by the Robust Vision Challenge rank.
}
\label{tab:comparison_rob}
\centering
\begin{adjustbox}{max width=\textwidth}
\begin{tabular}{l|cccc}
 & \multicolumn{1}{c}{~Middlebury~} & \multicolumn{1}{c}{~KITTI~} & \multicolumn{1}{c}{~MPI Sintel~} & \multicolumn{1}{c}{~HD1K~} \\ \hline
PWC-Net\_ROB~\cite{Sun2017} & 2 & 4 & 2 & 1 \\
ProFlow\_ROB~\cite{Maurer2018}& 1 & 6 & 1 & 4 \\
\textbf{ContinualFlow\_ROB} & 5 & 2 & 3 & 3 \\
LFNet\_ROB {[}anon{]}& 7 & 1 & 6 & 5 \\
AugFNG\_ROB {[}anon{]} & 9 & 3 & 4 & 2 \\
FF++\_ROB~\cite{Schuster2018a} & 3 & 5 & 5 & 6 \\
DMF\_ROB~\cite{Weinzaepfel2013} & 4 & 8 & 7 & 8 \\
ResPWCR\_ROB {[}anon{]} & 6 & 7 & 8 & 7 \\
WOLF\_ROB {[}anon{]} & 8 & 9 & 9 & 9 \\
TVL1\_ROB~\cite{Perez2013} & 10 & 10 & 10 & 10 \\
H+S\_ROB~\cite{Horn1981} & 11 & 11 & 11 & 11
\end{tabular}
\end{adjustbox}
\end{table}


\section{Conclusion}
The ContinualFlow network for optical flow estimation was introduced, with two novelties - occlusion estimation integrated in the optic flow computation  and the use of the optic flow from the previous time instant, and, through recursion, of all prior flows.
We showed that the two contributions improve performance, especially in occluded areas or areas close to motion discontinuities. 
In evaluation on standard dataset ContinualFlow is top ranked in Sintel and 3rd in KITTI.

\section*{Acknowledgements}
The research was supported by Toyota Motor Europe, CTU student grant \\
SGS17/185/OHK3/3T/13 and the OP VVV MEYS project\\
CZ.02.1.01/0.0/0.0/16\_019/0000765 Research Center for Informatics.


\bibliographystyle{splncs04}
\bibliography{jabref}

\end{document}